\documentclass[preprint,12pt]{elsarticle}

\usepackage{amssymb}
\usepackage{amsmath}
\usepackage{makecell}
\usepackage{booktabs}
\usepackage{xurl}

\begin{document}

\begin{frontmatter}



\title{Embedding -based Crop Type Classification in the Groundnut Basin of Senegal}


\author{Madeline C. Lisaius, Srinivasan Keshav, Andrew Blake} 

\affiliation{organization={The University of Cambridge Department of Computer Science and Technology},
            }

\author{Clement Atzberger} 

\affiliation{organization={dClimate Labs},
            }

\begin{abstract}
Crop type maps from satellite remote sensing are important tools for food security, local livelihood support and climate change mitigation in smallholder regions of the world, but most satellite-based methods are not well suited to smallholder conditions. To address this gap, we establish a four-part criteria for a useful embedding-based approach consisting of 1) performance, 2) plausibility, 3) transferability and 4) accessibility and evaluate geospatial foundation model (FM) embeddings -based approaches using TESSERA and AlphaEarth against current baseline methods for a region in the groundnut basin of Senegal. We find that the TESSERA -based approach to land cover and crop type mapping fulfills the selection criteria best, and in one temporal transfer example shows 28\% higher accuracy compared to the next best method. These results indicate that TESSERA embeddings are an effective approach for crop type classification and mapping tasks in Senegal.
\end{abstract}




\end{frontmatter}



\section{Introduction}
Reliable, wall-to-wall mapping of smallholder crops landscapes is an important tool for food security, climate adaptation, and sustainable land management globally.  Approximately three-quarters of the world’s poor live in rural areas and rely primarily on small-scale agriculture \cite{akpan_rural_2023}. These farms are critical not only for rural livelihoods but also for local and regional food security; globally, farms under 2 hectares (ha) contribute an estimated 30–34\% of food supply, devote a greater proportion of their production to food, and maintain greater crop diversity than larger farms \cite{ricciardi_how_2018}. Knowing what crops are grown where allows for informed decision making at regional, national and global scales that can mean survival for vulnerable people.  

In west Africa, much of the agricultural landscape is dominated by smallholder systems, where small plots support the livelihoods of most rural households, yet these agricultural systems remain understudied relative to large farms in the Global North, making crop mapping more difficult \cite{kamara_relevance_2019}. 

Satellite remote sensing provides an opportunity to bridge research and mapping gaps by enabling consistent, large-scale, and low-cost monitoring of agricultural landscapes. Freely available sensor data such as from Sentinel-1 and Sentinel-2 offer regularly repeating observations that can be leveraged for crop type mapping \cite{becker-reshef_crop_2023}. However, most existing methods have been developed in contexts of industrial agriculture, where fields are large, monocropped, and well-documented, making them poorly suited for smallholder regions \cite{vajsova_assessing_2020}. In West Africa, the challenges of frequent cloud cover, limited ground samples, small and irregular field sizes, and intercropping further complicate the creation of accurate, wall-to-wall crop type maps \cite{ibrahim_mapping_2021, kerner_how_2024}.

Recent advances in geospatial foundation models (FMs) offer a promising new approach. Embeddings derived from large-scale pre-trained models have shown strong performance on agricultural classification tasks, often matching or surpassing traditional machine learning approaches in accuracy and F1 score \cite{brown_alphaearth_2025, feng_tessera_2025, lisaius_using_2024}.  Yet, these methods have not been systematically evaluated under true smallholder conditions such as diverse management and intercropping which is common in West Africa.

To address this gap, we evaluate the potential of embedding-based approaches for wall-to-wall crop cover classification in the groundnut basin of Senegal, a region representative of many smallholder agricultural systems in west Africa. We compare multiple crop type classification approaches across several years and assess their performance and practical utility for real-world agricultural mapping applications.

\section{Recent work}

Some recent literature exists for wall-to-wall crop type mapping of agriculture in West Africa:

Ibrahim et al. \cite{ibrahim_mapping_2021} employ Sentinel-1, Sentinel-2 and SkySat imagery for land and crop cover mapping in a smallholder, intercropped region of Nigeria where average field size ranges between 0.3 ha and 0.5 ha. They first create a wall-to-wall classification map of land cover using spectral temporal metrics (STMs) from both Sentinel-1 and Sentinel-2, informed by cropping season. They then use the the land cover map to mask for agriculture regions and classify their five agricultural crops of interest using Random Forest on the STMs. They use high resolution images from SkySat to verify. This work requires extensive crop-specific feature engineering to prepare data for use by RF, therefore requiring customization to adapt to new regions.

Azzari et al. \cite{azzari_understanding_2021} use harmonic regressions to classify maize from non-maize in Malawi and Ethiopia using Sentinel-1 and Sentinel-2 images. Field sizes are not reported but study areas are noted as smallholder regions. Pixel-wise harmonic regressions along with rasterized weather data and topographic data are used as inputs to Random Forest classifier to distinguish maize and non-maize. This method relies on a high quality cropland mask to be applied beforehand, and relies on generation of the harmonic regression features for each year and region, and addition of other data such as topography and climate data. Other researchers using a similar harmonics approach also use it for few-class classification, which is unsuitable for mapping large, diverse crop regions and minority crops (Luan Pott 2022).

Gumma et al. \cite{gumma_dryland_2024} classify nine classes of crops in Senegal for 2020 using Sentinel-2 data, across all field sizes. They compare Random Forest, CART, SVM, and spectral matching technique (SMT). They prepare NDVI for the time frame of interest, group the NDVI time series with k-means clustering, and use labeled data to create ``ideal spectra" that are leveraged to classify the clusters by similarity to the ideal. They begin by classifying 10 land cover classes and 9 crop cover classes for all of Senegal. They find the spectral matching technique has the highest accuracy of the methods evaluated but requires extensive feature preparation.

Rustowicz et al. \cite{rustowicz_semantic_2019} approach crop type classification and segmentation in Ghana, South Sudan and Germany. They compare a 3D U-Net and the 2D convolutional neural network (CNN) with convolutional long short term memory model (CLSTM) that incorporates both CNNs and recurrent neural networks (RNNs) for semantic segmentation of multi-temporal, multi-spatial satellite images from Sentinel-2. They find that their proposed approach has the highest accuracy of tested methods in Germany. However for the smallholder regions they find that Random Forest often matches the accuracy of their approach. While this method is focused primarily on segmentation, it does consider the smallholder context. It is challenged by high cloud coverage as well as limited and imbalanced labels.


\section{Methods}
\subsection{Criteria}\label{sec:criteria}
To identify the best method for crop type mapping in a smallholder region like the groundnut basin of Senegal, we identify four qualities of a suitable wall-to-wall mapping approach:
\begin{enumerate}
      \item \textbf{Performance:} A good approach must offer high accuracy and F1 scores in crop type classification. As high quality datasets are difficult to acquire and maintain in smallholder regions, we acknowledge that these metrics may be lower than those in regions with high data quality and availability.

    \item \textbf{Plausibility:} A good approach will show mapping results that are plausible for the region of study. Maps across years should show consistency, and additionally, changes to the extent of cropland should match trends in cropland statistics.  

    \item \textbf{Transferability:} As many organizations concerned with smallholder agriculture in the Global South do not have the capacity or resources to collect ground truth labels every year, it's important that a modeling approach trained on one year can be transferred to other years. The best approach will maintain accuracy when transferring between years, taking into account that limited labels and varying label quality will impact accuracy. 

    \item \textbf{Accessibility:} Given that projects supporting smallholder and subsistence agriculture often have fewer resources than industrial agriculture, approaches that eliminate the need for hand-engineered features and have reduced computational resource requirements are preferred to make mapping more accessible.
\end{enumerate}

\subsection{Labeled data preparation} \label{sec:landcoverdataprep}
Data are obtained from the Joint Experiment for Crop Assessment and Monitoring (JECAM) initiative for a region of Senegal spanning the Fatick and Niakhar departments \cite{jolivot_harmonized_2023}. Fatick and Niakhar lie in the southeast of Senegal's groundnut basin and primarily produce groundnuts and millet in biennial rotation as well as livestock \cite{jolivot_harmonized_2023}. The region is challenged by water scarcity, exacerbated by climate change, food insecurity in over 80\% of the population, and the combination of population increase and natural resource degradation \cite{piraux_exploring_2023}.   

\begin{figure}[h!]
  \centering
  \includegraphics[width=350pt]{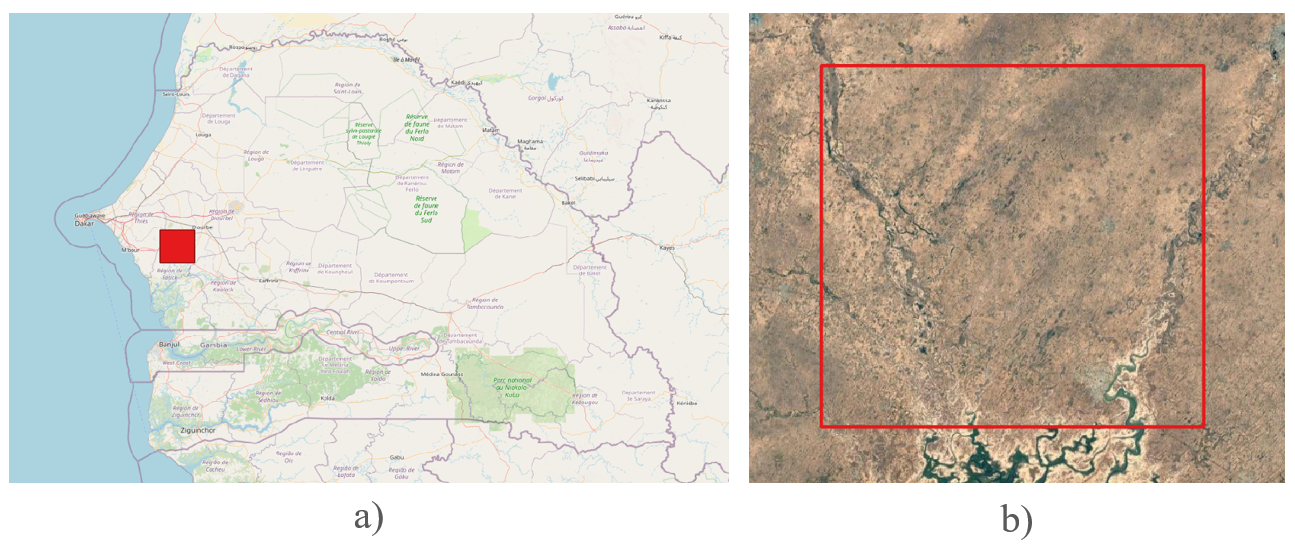}
  \caption{Visualized are a) the location of the region of interest within Senegal and b) Sentinel-2 satellite imagery of the region of interest. This region is dominated by dry shrub land that has been converted into agricultural land and features a river delta in the south.}
  \label{fig:senegalmap}
\end{figure}

The JECAM ground truth data consists of 734 labeled polygons in 2018, 669 labeled polygons in 2019, and 1870 labeled polygons in 2021. Waypoints were collected with a stratified strategy across the area of interest, and for each a land cover classification is given. For agricultural waypoints, a primary crop is indicated, and in cases of intercropping, a secondary crop is labeled as well.  Further details can be found in the paper \cite{jolivot_harmonized_2023}.

The 2021 polygons have additional classes, both for land cover and for crop types. We standardize the land cover classification across years by removing the two additional classes,``pasture" and ``natural vegetation", which had limited samples and significant overlap with the existing ``shrub land" class for 2021. Data cleaning was also required to consolidate crops that were classified under multiple names, to group classes with very few labels (fewer than 10 examples per class) and to harmonize with the 2018 and 2019 data. For example, the labels ``bissap" and ``hibiscus" are combined as they reference the same crop, and are afterwards grouped into the ``other" label. 

Classes and their counts for each year are presented in Table \ref{tab:landcoverpolygons} for land cover and Table \ref{tab:cropcoverpolygons} for crop type. The average size of fields for each crop class is shown in Table \ref{tab:cropcoverareas}, with the highest average field size (excluding fallow) being 0.95 ha for the tree class in 2021, and the smallest being 0.24 ha for cowpea in 2021. These areas are well below the global threshold of small farms of 2 ha \cite{lowder_number_2016}. Additionally, 94.3\% of labeled fields in 2018, 93.7\% of labeled fields in 2019, and 92.3\% of labeled fields in 2021 are smaller than 2 ha, situating the study area as a smallholder region. The percentage of fields that are that are marked as intercropped is shown in Table \ref{tab:intercropping}, grouped by the primary crop.

\begin{table}[h!]
\centering
\footnotesize
\caption{\textbf{Number of polygons per land cover class, per year}. Note that the two additional classes in 2021 in \textit{italic} are not included in classification but are shown for illustrative purposes. 2021 shows many more crop land labels than 2018 and 2019.}
\label{tab:landcoverpolygons}
\resizebox{\textwidth}{!}{%
\begin{tabular}{lccccccccc}
\hline
Year & Total & Crop & Water & Shrub & Bare & Wetland & Built up & Natural & Pasture \\ 
& polygons & land & body & land & soil & & surface & vegetation \\
\hline
2018 & 734  & 551 & 55  & 50  & 35  & 22  & 21 & - & \\
2019 & 669  & 486 & 55  & 50  & 35  & 22  & 21 & - & \\
2021 & 1870 & 1454 & 82 & 123 & 44  & 31  & 45 & \textit{87} & \textit{4} \\
\hline
\end{tabular}%
}
\end{table}

\begin{table}[h!]
\centering
\footnotesize
\caption{The \textbf{number of polygons per crop type, per year }varies between years, especially in 2021 where the proportion of groundnut and millet are much higher. The 2021 labeled dataset has additional classes which are grouped into `rice', `tree' and `other.'}
\label{tab:cropcoverpolygons}
\begin{tabular}{lcccccccccc}
\hline
Year & Cowpea & Fallow & Groundnut & Millet & Sorghum & Rice & Tree & Other\\
\hline
2018 & 24  & 69  & 198 & 225 & 35 & -  & -  & -  \\
2019 & 14  & 27  & 134 & 259 & 52 & -  & -  & -  \\
2021 & 45  & 79  & 467 & 608 & 92 & 14 & 45 & 98 \\
\hline
\end{tabular}
\end{table}

\begin{table}[h!]
\centering
\footnotesize
\caption{The \textbf{average field area in ha, per crop type and per year}, are smaller than the typical 2 ha area for a ``smallholder" field size.}
\label{tab:cropcoverareas}
\begin{tabular}{lcccccccc}
\hline
Year & Cowpea & Fallow & Groundnut & Millet & Sorghum & Rice & Tree & Other \\
\hline
2018 & 0.29 & 1.24 & 0.44 & 0.45 & 0.34 & - & - & - \\
2019 & 0.34 & 0.22 & 0.72 & 0.52 & 0.48 & - & - & - \\
2021 & 0.24 & 1.11 & 0.52 & 0.52 & 0.52 & 0.92 & 0.95 & 0.74 \\
\hline
\end{tabular}
\end{table}

\begin{table}[h!]
\centering
\footnotesize
\caption{The \textbf{percent of fields for each main crop that are reported intercropped} varies considerably by year. In particular, almost no fields from 2019 are reported as intercropped, which is a meaningful difference from 2018 and 2021.}
\label{tab:intercropping}
\begin{tabular}{lcccccccccc}
\hline
Year & Cowpea & Fallow & Groundnut & Millet & Sorghum & Rice & Tree & Other\\
\hline
2018 & 12.5\%  & 0\% & 77.8\% & 7.1\% & 20\% & -  & -  & -  \\
2019 & 0\%  & 0\%  & 0\% & 12.4\% & 0\% & -  & -  & -  \\
2021 & 4.4\%  & 5.1\%  & 11.1\% & 3.5\% & 15.3\% & 0\% & 17.4\% & 26.7\% \\
\hline
\end{tabular}
\end{table}

The distribution of labels across years is unbalanced, with proportions shifting considerably even among classes present in all years. For instance, in the land cover data (Table \ref{tab:landcoverpolygons}), cropland consistently dominates but its proportion relative to other classes changes. Meanwhile, classes such as water body, shrub land, and bare soil remain comparatively minor. A similar pattern appears in the crop type data (Table \ref{tab:cropcoverpolygons}), where millet and groundnut proportions vary between years. Table \ref{tab:intercropping} furthermore shows that the percentage of fields that are marked as intercropped also changes significantly by year - most fields observed in 2019 are not intercropped, where as up to 77\% of groundnut fields in 2018 are intercropped. Similarly, the average size of fields for each crop class changes meaningfully by year as shown in Table \ref{tab:cropcoverareas}. This suggests likely different label collection strategies between years.

To address class imbalance, we implement class weightings in training and, for applicable methods, we use focal loss \cite{lin_focal_2018}. We experimented with adding Synthetic Minority Over-sampling Technique, SMOTE, \cite{chawla_smote_2002} for data augmentation but did not find that it improved classification accuracy and F1 scores, and as a result this is not used in the work.

\begin{figure}[h!]
  \centering
  \includegraphics[width=350pt]{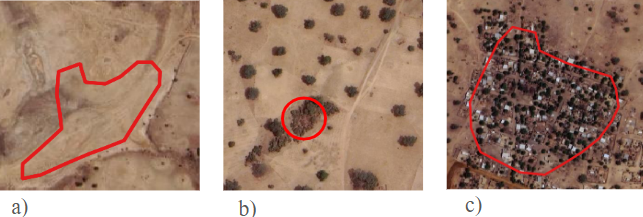}
  \caption{Pictured are three example label polygons, not at the same scale, showing a) bare soil, b) shrub land, and c) build-up surface. There is meaningful overlap in characteristics of ground cover amongst the three, particularly with build-up surface.}
  \label{fig:builtup}
\end{figure}

Additionally, there is meaningful overlap between classes in the dataset. For example, the classes ``bare soil" and ``shrub land" share characteristics with ``built-up" which is exemplified in Figure \ref{fig:builtup}. As exemplified in this case, assigning semantic labels to land cover is challenging, as human-defined categories often overlap in practice. The Senegal dataset illustrate the difficulty of mapping real-world heterogeneity into discrete labels, which is observed across label datasets in remote sensing \cite{hauser_perfect_2025}. While we note the challenges present in this dataset, we still acknowledge that this dataset is one of the best publicly available. Broadly, we see the pressing need for cleaner labels in the smallholder domain.

\subsection{Remote sensing data}
We use four data preparations as inputs: cloud-masked Sentinel-1 and Sentinel-2 time series data with vegetation indices (VIs) which will be referred to henceforth as ``raw", spectral-temporal metrics (STMs), TESSERA embeddings, and AlphaEarth embeddings. We choose to compare the FMs TESSERA and AlphaEarth because they are the only FMs currently provide pre-generated embeddings through free, publicly accessible platforms. We identify these as the best comparisons for crop type classification for the smallholder Senegal study region.

To prepare data for the ``raw'' and ``STM'' approaches, we collected Sentinel-1 and Sentinel-2 time series for each year of the study from 1 January to 31 December 2022 (excluding observations with 99\% cloud coverage or higher). From Sentinel 2, band numbers 2, 3, 4, 5, 6, 7, 8, 8a, 11, and 12 are used. All bands are normalized and resampled to 10 m spatial resolution. The QAI band is used for cloud masking. 

For the raw data input we compute five optical and SAR VIs commonly included: Normalized Difference Vegetation Index (NDVI), Green Chlorophyll Vegetation Index (GCVI), Enhanced Vegetation Index (EVI), Land Surface Water Index (LSWI) and the Sentinel-1 SAR -based Radar Vegetation Index (RVI). These are used with the Sentinel-1 observations and the Sentinel-2 observations with cloudy dates masked.

For preparation of STMs as defined in Ibrahim et al. \cite{ibrahim_mapping_2021}, we generate pixel-based Sentinel-2 metrics for six time windows: (i) all available dates, (ii) the beginning of season July 1st - August 15th, (iii) the peak season August 16th - October 15th, (iv) end of season October 16th - November 30th, (v) the full season July 1st - November 30th, and (vi) the off season December 1st through June 30th. We define the growing season by metadata in the JECAM dataset \cite{jolivot_harmonized_2023} and cross reference with FAO Senegal reports \cite{fao_state_2021}. For each temporal window, we produce metrics: minimum, 25\% quantile, 50\% quantile, 75\% quantile, maximum, average, standard deviation, range, inter-quartile range, skewness, kurtosis, NDVI, EVI, Normalized Difference Water Index (NDWI) and the Tasseled Cap transformations of wetness (TCW), greenness (TCG), and brightness (TCB) using Tasseled Cap values for Sentinel-2 from Jiang et al. \cite{jiang_derivation_2024}. We use these Sentinel-2 STMs and the Sentinel-1 time series with RVI added. We download annual TESSERA embeddings using the GeoTessera Python package for each year of the study \cite{cambridge_ucam-eogeotessera_2025}. TESSERA trains on global Sentinel-1 and Sentinel-2 imagery, learning via a sparse temporal augmentation approach to generate embeddings of 128 dimensions. 

We download annual Google AlphaEarth embeddings \cite{brown_alphaearth_2025} from Google Earth Engine in August 2025 as produced by v2.1 of the AlphaEarth Foundations model, which is an improvement on v2.0 evaluated in the published paper. This version detail was noted in the Earth Engine documentation at the time of download. AlphaEarth trains on multiple data modalities including optical (Sentinel-2, Landsat 8/9), radar (Sentinel-1, PALSAR2), LiDAR (GEDI), environmental (GLO-30, ERA5-Land, GRACE), and annotated (NLCD, Wikipedia) as well as spatial contextual information to generate a 64 -dimension annual embedding.

The number of features for each method is shown in Table \ref{tab:numfeatures}.

\begin{table}[ht]
\centering
\caption{The number of features of each data input is compared, showing that AlphaEarth has the lowest number of features, about 6\% of the raw approach.}
\begin{tabular}{lc}
\toprule
\hline
\textbf{Data input} & \textbf{Number of features} \\
\midrule
\hline
TESSERA & 128 \\
AlphaEarth & 64 \\
STM & 228 \\
Raw & 1106 \\
\bottomrule
\hline
\end{tabular}
\label{tab:numfeatures}
\end{table}

\subsection{Model training and evaluation}
For our approach, we first identify best classifiers, use these to ensemble, and then proceed with evaluation work.

In model training, we split labeled polygons into train, validation and test sets. As most polygons are collected in a spatially distributed manner across the study region, we believe this split automatically should account for issues of spatial autocorrelation. All pixels in each polygon are used. 

To identify the best classifier heads for each input data type, we train the classifiers Random Forest (RF), XGBoost, Support Vector Machine (SVM), Logistic Regression (LR) and Multi-Layer Perceptron (MLP) each 200 times with different seeds. The accuracy, macro F1 and weighted F1 scores of this single classifier evaluation are shown in Table \ref{tab:classification2018_singlemethods}. For ensembling, we take the two strongest classifiers for each data input; as even over 200 runs there is overlap between metrics, we take the two classification heads with the highest mean metrics. We also note that there is high standard deviation for most metrics, which reflects the instability of using a single model approach in this context.

For TESSERA we select MLP and XGBoost, for AlphaEarth we select MLP and LR, for raw we select MLP and XGBoost, and for STMs we select RF and XGBoost. In the ensemble approach, the class-wise predicted probabilities for each pixel are averaged across the two selected classifiers. We classify each pixel using the class with the highest overall probability. For the remainder of this chapter, each data input's ensemble will simply be referred to as ``ensemble", for example ``TESSERA ensemble."

\begin{table}[h!]

\centering

\caption{Classification results (2018, land cover) for the test set organized by method and metric across data inputs. Values of  $\pm$ are standard deviation over 200 runs. The two methods selected for ensembling for each input (by column) are in \textbf{bold}. As there is overlap between values, the methods are selected by highest mean value across all three metrics.}

\label{tab:classification2018_singlemethods}

\renewcommand{\arraystretch}{1.15}

\resizebox{\textwidth}{!}{%

\begin{tabular}{llcccc}

\hline

\textbf{Method} & \textbf{Metric} & \textbf{AlphaEarth} & \textbf{TESSERA} & \textbf{Raw} & \textbf{STM} \\

\hline

\small

MLP & Accuracy   & \textbf{0.936 $\pm$ 0.019} & \textbf{0.944 $\pm$ 0.016} & \textbf{0.933 $\pm$ 0.015} & 0.835 $\pm$ 0.030 \\

    & Macro F1   & \textbf{0.872 $\pm$ 0.043} & \textbf{0.884 $\pm$ 0.037} & \textbf{0.859 $\pm$ 0.041} & 0.565 $\pm$ 0.104 \\

    & Weighted F1& \textbf{0.934 $\pm$ 0.021} & \textbf{0.945 $\pm$ 0.015} & \textbf{0.932 $\pm$ 0.015} & 0.808 $\pm$ 0.040 \\

\hline

Random        & Accuracy   & 0.918 $\pm$ 0.027 & 0.935 $\pm$ 0.015 & 0.919 $\pm$ 0.019 & \textbf{0.912 $\pm$ 0.023} \\

Forest        & Macro F1   & 0.828 $\pm$ 0.048 & 0.866 $\pm$ 0.040 & 0.829 $\pm$ 0.050 & \textbf{0.796 $\pm$ 0.068} \\

              & Weighted F1& 0.911 $\pm$ 0.033 & 0.934 $\pm$ 0.015 & 0.917 $\pm$ 0.020 & \textbf{0.930 $\pm$ 0.025} \\

\hline

Logistic         & Accuracy   & \textbf{0.935 $\pm$ 0.014} & 0.933 $\pm$ 0.016 & 0.923 $\pm$ 0.010 & 0.891 $\pm$ 0.021 \\

Regression       & Macro F1   & \textbf{0.865 $\pm$ 0.033} & 0.850 $\pm$ 0.035 & 0.844 $\pm$ 0.030 & 0.781 $\pm$ 0.049 \\

                       & Weighted F1& \textbf{0.934 $\pm$ 0.015} & 0.933 $\pm$ 0.016 & 0.923 $\pm$ 0.010 & 0.919 $\pm$ 0.023 \\

\hline

SVM & Accuracy   & 0.931 $\pm$ 0.006 & 0.930 $\pm$ 0.016 & 0.876 $\pm$ 0.017 & 0.876 $\pm$ 0.030 \\

    & Macro F1   & 0.864 $\pm$ 0.018 & 0.844 $\pm$ 0.029 & 0.799 $\pm$ 0.024 & 0.780 $\pm$ 0.041 \\

    & Weighted F1& 0.932 $\pm$ 0.006 & 0.931 $\pm$ 0.015 & 0.882 $\pm$ 0.019 & 0.882 $\pm$ 0.029 \\

\hline

XGBoost & Accuracy   & 0.922 $\pm$ 0.029 & \textbf{0.941 $\pm$ 0.017} & \textbf{0.928 $\pm$ 0.015} & \textbf{0.908 $\pm$ 0.023} \\

        & Macro F1   & 0.850 $\pm$ 0.045 & \textbf{0.877 $\pm$ 0.039} & \textbf{0.832 $\pm$ 0.034} & \textbf{0.809 $\pm$ 0.047} \\

        & Weighted F1& 0.917 $\pm$ 0.034 & \textbf{0.941 $\pm$ 0.016} & \textbf{0.926 $\pm$ 0.016} & \textbf{0.906 $\pm$ 0.024} \\

\hline

\end{tabular}%

}

\end{table}

To evaluate the ensemble approaches, we train, validate and evaluate the ensemble 20 times with different seeds used for model state and data split, and report the accuracy and F1 scores. This is used for evaluation criterion 1, Performance. To generate the full land cover map, we use all the labeled data to train and validate with a 90/10 split, aggregating probabilities from three training runs with different random seeds, and use this to complete the wall-to-wall classification.

Given the limitations of the labeled data, we also visually inspect the wall-to-wall maps and consider change over time.  This step pertains to evaluation Criterion 2, Plausibility. To consider Criterion 3, Transferability, we train the land cover ensemble methods on 2018 and evaluate classification on 2019 and 2021, and train on 2019 and test on 2018 and 2021. To consider Criterion 4, Accessibility, we consider CPU used by each method and requirements of task-specific feature engineering. 

We proceed using ``core cropland'', which we define as the area that is classified consistently as cropland across the three years, to mask for cropland before classifying the crop type. Crops are considered only with the primary label regardless of intercropping. We consider the accuracy and F1 scores for evaluation Criterion 1, Performance, and we inspect wall-to-wall maps for Criterion 2, Plausibility.

\section{Results}

\subsection{Ensemble land cover classification}\label{sec:ensembleclassification}

For all years of study, the TESSERA -based ensembles have comparable or higher accuracy and F1 scores for land cover classification than the competing methods as shown in Table \ref{tab:ensemble_results}. In considering evaluation Criterion 1, Performance, across all years of results, the TESSERA embedding -based approach and the AlphaEarth embedding -based approach are comparable across accuracy and F1 scores but AlphaEarth has a much higher standard deviation across runs, suggesting that these embeddings offer less stable classification than that of TESSERA. The raw data approach also performs comparably to TESSERA and AlphaEarth in 2018.

\begin{table}[h!]
\centering
\footnotesize
\renewcommand{\arraystretch}{1.2}
\caption{\textbf{Land cover} classification performance across years and ensemble methods on the withheld test set. Best approaches by year are in \textbf{bold}. Values of $\pm$ denote standard deviation over 20 runs. Across all years, TESSERA and AlphaEarth have the highest performance metrics.}
\label{tab:ensemble_results}
\begin{tabular}{llccc}
\hline
\textbf{Year} & \textbf{Ensemble Method} & 
\makecell{\textbf{Accuracy} \\ (mean $\pm$ std)} & 
\makecell{\textbf{Macro avg F1} \\ (mean $\pm$ std)} & 
\makecell{\textbf{Weighted avg F1} \\ (mean $\pm$ std)} \\
\hline
2018 & Raw         & 0.946 $\pm$ 0.012 & 0.847 $\pm$ 0.014 & 0.944 $\pm$ 0.013 \\
     & STM         & 0.933 $\pm$ 0.009 & 0.828 $\pm$ 0.005 & 0.930 $\pm$ 0.008 \\
     & TESSERA     & \textbf{0.965 $\pm$ 0.007} & \textbf{0.906 $\pm$ 0.009} & \textbf{0.965 $\pm$ 0.012} \\
     & AlphaEarth  & \textbf{0.956 $\pm$ 0.019} & \textbf{0.887 $\pm$ 0.013} & \textbf{0.953 $\pm$ 0.015} \\
\hline
2019 & Raw         & \textbf{0.957 $\pm$ 0.018} & \textbf{0.892 $\pm$ 0.017} & \textbf{0.956 $\pm$ 0.019} \\
     & STM         & 0.920 $\pm$ 0.016 & 0.849 $\pm$ 0.012 & 0.917 $\pm$ 0.015 \\
     & TESSERA     & \textbf{0.967 $\pm$ 0.008} & \textbf{0.903 $\pm$ 0.012} & \textbf{0.962 $\pm$ 0.005} \\
     & AlphaEarth  & \textbf{0.944 $\pm$ 0.019} & \textbf{0.832 $\pm$ 0.018} & \textbf{0.941 $\pm$ 0.012} \\
\hline
2021 & Raw         & 0.872 $\pm$ 0.010 & 0.708 $\pm$ 0.009 & 0.862 $\pm$ 0.014 \\
     & STM         & 0.855 $\pm$ 0.022 & 0.642 $\pm$ 0.018 & 0.840 $\pm$ 0.015 \\
     & TESSERA     & \textbf{0.911 $\pm$ 0.011} & \textbf{0.775 $\pm$ 0.016} & \textbf{0.907 $\pm$ 0.012} \\
     & AlphaEarth  & \textbf{0.907 $\pm$ 0.013} & \textbf{0.742 $\pm$ 0.024} & \textbf{0.901 $\pm$ 0.015} \\
\hline
\end{tabular}
\end{table}

\subsection{Wall-to-wall land cover maps}

\begin{figure}[h!]
  \centering
  \caption{Land cover maps}
  \includegraphics[width=370pt]{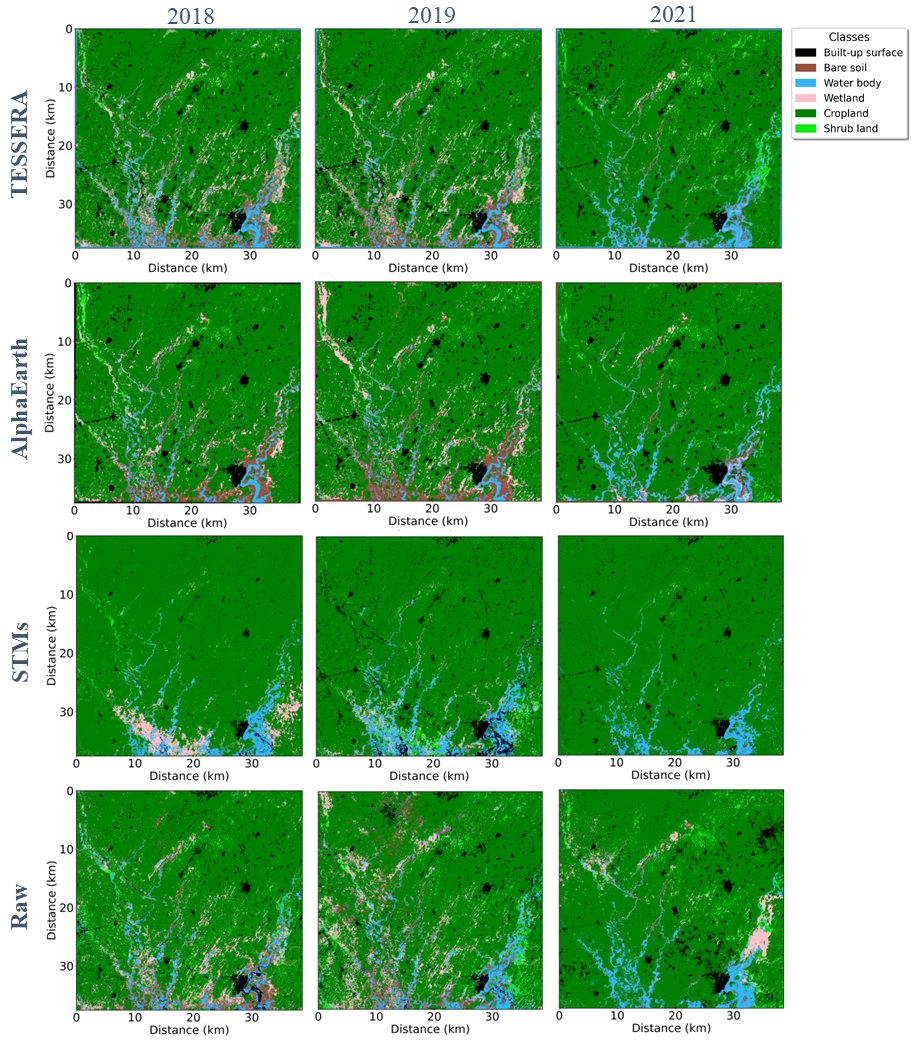}
  \caption{Pictured are the land cover ensemble maps using TESSERA, AlphaEarth, STMs, and raw input data in 2018, 2019 and 2021. Units on the x and y axis are kilometers of distance. Across all methods, the maps for 2021 are noticeably different, suggesting a differing quality of labeled data in those years.}
  \label{fig:all_landcover}
\end{figure}

 Wall-to-wall land cover classification maps can be seen for all data approaches in Figure \ref{fig:all_landcover}. For all approaches, maps for 2021 noticeably shift, likely reflecting a change to the quality of labels in 2021. Across the three years, the TESSERA -based approach and AlphaEarth -based approach show the most consistency, which pertains to Criterion 2, Plausibility. Despite the fact that both TESSERA and AlphaEarth approaches have high classification accuracy and F1 scores and both show consistency between years, there are differences between the two approaches' maps. This may be a result of the limited training data which does not represent the full range of conditions present in the actual landscape interacting with the encoded information of input embeddings, and further emphasizes the need for high quality, semantically coherent labeled data

\subsection{Temporal transfer learning for land cover classification}

When training on 2018 and evaluating on 2019 and 2021, and when traing on 2019 and evaluating on 2018 and 2021, TESSERA, AlphaEarth and STM -based approaches largely overlap in accuracy and F1 scores for land cover classification. These methods come close to and even surpass the best accuracy for same-year training, $0.965 \pm 0.007$ with TESSERA and $0.956 \pm 0.019$ for AlphaEarth, although in this transfer learning case, the approaches are trained on more labels due to the lack of necessity to withhold an evaluation set. This suggests that land cover classification is a relatively stable task in this region and is transferable across years.

\begin{table}[h!]
\centering
\caption{Cross-year land classification performance by training year, prediction year, and method. Values of  $\pm$ are standard deviation over 5 runs and are from differences in the validation set used. }
\label{tab:crossyear_results}
\footnotesize
\begin{tabular}{cc l ccc}
\hline
Method & \makecell{Training \\ year} & \makecell{Prediction \\ year} & 
\makecell{Accuracy \\ (mean $\pm$ std)} & 
\makecell{Macro avg F1 \\ (mean $\pm$ std)} & 
\makecell{Weighted avg F1 \\ (mean $\pm$ std)} \\
\hline
TESSERA & 2018 & 2019  & 0.958 $\pm$ 0.002 & 0.903 $\pm$ 0.003 & 0.957 $\pm$ 0.002 \\
& & 2021 & 0.850 $\pm$ 0.003 & 0.760 $\pm$ 0.002 & 0.846 $\pm$ 0.003 \\
& 2019 & 2018  & 0.961 $\pm$ 0.001 & 0.907 $\pm$ 0.002 & 0.961 $\pm$ 0.001 \\
& & 2021 & 0.847 $\pm$ 0.002 & 0.691 $\pm$ 0.003 & 0.856 $\pm$ 0.002 \\
\hline
AlphaEarth & 2018 & 2019 & 0.941 $\pm$ 0.002 & 0.852 $\pm$ 0.003 & 0.936 $\pm$ 0.002 \\ 
& & 2021 & 0.844 $\pm$ 0.003 & 0.610 $\pm$ 0.002 & 0.839 $\pm$ 0.003 \\
& 2019 & 2018  & 0.967 $\pm$ 0.001 & 0.914 $\pm$ 0.002 & 0.968 $\pm$ 0.002 \\
& & 2021 & 0.851 $\pm$ 0.002 & 0.660 $\pm$ 0.003 & 0.852 $\pm$ 0.002 \\
\hline
STM & 2018 & 2019 & 0.942 $\pm$ 0.002 & 0.900 $\pm$ 0.003 & 0.941 $\pm$ 0.002 \\
& & 2021 & 0.837 $\pm$ 0.003 & 0.682 $\pm$ 0.002 & 0.836 $\pm$ 0.003 \\
& 2019 & 2018  & 0.973 $\pm$ 0.001 & 0.926 $\pm$ 0.002 & 0.973 $\pm$ 0.001 \\
& & 2021 & 0.843 $\pm$ 0.002 & 0.694 $\pm$ 0.003 & 0.847 $\pm$ 0.002 \\
\hline
Raw & 2018 & 2019  & 0.814 $\pm$ 0.003 & 0.552 $\pm$ 0.002 & 0.798 $\pm$ 0.003 \\ 
& & 2021 & 0.830 $\pm$ 0.003 & 0.720 $\pm$ 0.003 & 0.830 $\pm$ 0.002 \\
& 2019 & 2018 & 0.953 $\pm$ 0.002 & 0.882 $\pm$ 0.003 & 0.951 $\pm$ 0.002 \\ 
& & 2021 & 0.832 $\pm$ 0.003 & 0.700 $\pm$ 0.002 & 0.833 $\pm$ 0.002 \\
\hline
\end{tabular}
\end{table}

\subsection{CPU use}

For land cover classification, AlphaEarth and TESSERA -based classification methods are computationally leaner than the raw and STM -based approaches, as shown in Table \ref{tab:cpuuse}. Additionally, the pre-computed TESSERA and AlphaEarth -based approaches require no feature engineering such as computing VI, NDVI composites, or STM generation, reducing further CPU demand. They also necessitate no specialized knowledge about cropping seasons, meaningfully simplifying the classification process. Considering both CPU use and feature engineering, AlphaEarth and TESSERA lead for Criterion 4, Accessibility. 

\begin{table}[ht]
\centering
\caption{Relative CPU performance ratios across methods for ensemble land cover classification are compared, with AlphaEarth requiring the lowest amount of compute. Confidence intervals are shown with $\pm$ from 20 runs. }
\begin{tabular}{lc}
\hline
\toprule
Ensemble method & Relative CPU use \\
\midrule
\hline
TESSERA & 1 \\
STM ensemble & 6.99$\pm$0.51 \\
Raw ensemble & 4.60$\pm$0.26 \\
AlphaEarth ensemble & 0.67$\pm$ 0.13 \\
\bottomrule
\hline
\end{tabular}
\label{tab:cpuuse}
\end{table}

\subsection{Crop land change across years}\label{sec:changeacrossyear}

The district of Fatick is estimated to have approximately 2\% of land in urban or build-up spaces and 56\% agricultural land \cite{piraux_exploring_2023}. The subregion of this study is the most agriculturally intensive of Fatick and consequentially we expect the value for agricultural land to be considerably higher. We quantify the area of core cropland and we expect the value to be above 56\%. We additionally compare the areas that change to or away from cropland between each pair of years 2018-2019, 2018-2021 and 2019-2021, as shown in Table \ref{tab:change_results}. We expect no more than 2-3\% change to agriculture land each year as noted in Section \ref{sec:changeacrossyear}. Besides this, we also would expect a slight decline in the agricultural area in 2019 relative to 2018, given reports of decreased farm cultivation areas in the Fatick and Niakhar regions \cite{piraux_exploring_2023}.

In comparing cropland areas identified between years, the TESSERA and STM -based approaches have the most realistic estimates of change, with 3\% or lower overall change per year and reflecting a slight decrease in cropland area in 2019 as expected. Meanwhile, AlphaEarth and raw -based methods show much higher (6-11\%) and less plausible change between years. This suggests that TESSERA and STM -based methods are more stable across years, related to criterion 2, Plausibility.

\begin{table}[h!]
\centering
\caption{Percentage changes of wall-to-wall measured cropland between years for different inputs. Raw inputs result in the highest measured change in crop land area between years (an increase in Cropland between 2019-2021 of 15\%) where as TESSERA and STM inputs result in the lowest.}
\label{tab:change_results}
\footnotesize
\begin{tabular}{llccc}
\hline
Dataset & Year Comparison & Decrease (\%) & Increase (\%) & Aggregate Change (\%) \\
\hline
Raw        & 2018--2019 & 10.7 & 4.7 & 15.4 \\
           & 2018--2021 & 5.8 & 10.7 & 16.5 \\
           & 2019--2021 & 4.7 & 15.8 & 20.5 \\
\hline
TESSERA    & 2018--2019 & 1.1 & 1.0 & 2.1 \\
           & 2018--2021 & 1.2 & 2.0 & 3.2 \\
           & 2019--2021 & 1.0 & 2.0 & 3.0 \\
\hline
STM        & 2018--2019 & 0.9 & 2.6 & 3.5 \\
           & 2018--2021 & 1.7 & 1.6 & 3.3 \\
           & 2019--2021 & 2.7 & 0.9 & 3.6 \\
\hline
AlphaEarth & 2018--2019 & 4.8 & 1.6 & 6.4 \\  
           & 2018--2021 & 2.1 & 6.4 & 8.5 \\ 
           & 2019--2021 & 2.3 & 8.8 & 11.1 \\ 
\hline
\end{tabular}
\end{table}

Additionally, we observe that the area of the core cropland determined by ensemble method varies from ~64\% for raw input to ~81\% for STM as shown in Table \ref{tab:corecropland}. However this range is within the expected percentage.

\begin{table}[ht]
\centering
\caption{Percent of area as core cropland which is calculated by comparing the area classified as cropland consistently across all three years of data: 2018, 2019 and 2021.}
\begin{tabular}{lc}
\hline
\toprule
Method & Percent cropland \\
\midrule
\hline
TESSERA ensemble & 70.6 \\
STM ensemble & 81.1 \\
Raw ensemble & 61.6 \\
AlphaEarth ensemble & 73.5 \\
\bottomrule
\hline
\end{tabular}
\label{tab:corecropland}
\end{table}

\subsection{Ensemble crop cover classification}

For all years, the TESSERA -based ensemble for crop type classification has comparable or higher accuracy and F1 scores than competing methods, as shown in Table \ref{tab:ensemble_results_crop}. It is notable that crop type classification accuracy decreases each year, with an almost 30\% difference between the 2018 and 2021 values. As discussed in Section \ref{sec:landcoverdataprep} and as seen in Table \ref{tab:cropcoverpolygons} and Table \ref{tab:intercropping}, there are meaningful differences in the number and types of of samples collected between years and evidence for changing data collection methods which may explain this change. Land cover classification results (Table \ref{tab:ensemble_results_crop}) show little difference in land classification accuracy between 2018 and 2019, with notably poorer classification in 2021. This suggests that labels from 2019 and 2021 are of lower quality. This further emphasizes the importance of trustworthy labeled data for crop type mapping. Additionally, the TESSERA embedding -based approach maintains 4\% to 8\% advantage in every metric over AlphaEarth. This contrasts with land cover classification where AlphaEarth embeddings performed comparably to TESSERA. Additionally, the AlphaEarth embedding -based approach performs worse in some cases than both the STM and raw approaches. This suggests that embedding -based approaches are not always a better base for classification tasks and, as exemplified by AlphaEarth, some embeddings may be better suited to some classification tasks more than others.

\begin{table}[h!]
\centering
\footnotesize
\caption{\textbf{Crop cover} classification performance across years and ensemble methods is reported with best values per year in \textbf{bold}. In all cases, the models are trained and evaluated on the same year. Values of  $\pm$ are standard deviation over 20 runs. Across all three years, TESSERA results in higher mean performance metrics.}
\label{tab:ensemble_results_crop}
\begin{tabular}{llccc}
\hline
Year & Ensemble Method & 
\makecell{Accuracy \\ (mean $\pm$ std)} & 
\makecell{Macro avg F1 \\ (mean $\pm$ std)} & 
\makecell{Weighted avg F1 \\ (mean $\pm$ std)} \\
\hline
2018 & Raw        & 0.786 $\pm$ 0.034 & 0.616 $\pm$ 0.044 & 0.781 $\pm$ 0.035 \\
 (5 crops)& STM        & 0.758 $\pm$ 0.040 & 0.524 $\pm$ 0.027 & 0.742 $\pm$ 0.039 \\
 & TESSERA    & \textbf{0.846} $\pm$ 0.033 & \textbf{0.643} $\pm$ 0.029 & \textbf{0.837} $\pm$ 0.035 \\
 & AlphaEarth & 0.784 $\pm$ 0.026 & 0.596 $\pm$ 0.042 & 0.778 $\pm$ 0.028 \\
\hline
2019 & Raw        & 0.647 $\pm$ 0.039 & 0.386 $\pm$ 0.056 & 0.61 $\pm$ 0.038 \\
 (5 crops)& STM        & 0.621 $\pm$ 0.060 & 0.362 $\pm$ 0.036 & 0.578 $\pm$ 0.065 \\
 & TESSERA    & \textbf{0.694} $\pm$ 0.045 & \textbf{0.420} $\pm$ 0.039 & \textbf{0.661} $\pm$ 0.044 \\
 & AlphaEarth & 0.575 $\pm$ 0.078 & 0.362 $\pm$ 0.061 & 0.558 $\pm$ 0.078 \\
\hline
2021 & Raw        & 0.516 $\pm$ 0.029 & 0.381 $\pm$ 0.042 & 0.517 $\pm$ 0.032 \\
 (8 crops)& STM        & 0.504 $\pm$ 0.024 & 0.339 $\pm$ 0.040 & 0.480 $\pm$ 0.030 \\
 & TESSERA    & \textbf{0.567} $\pm$ 0.036 & \textbf{0.418} $\pm$ 0.024 & \textbf{0.554} $\pm$ 0.038 \\
 & AlphaEarth & 0.490 $\pm$ 0.025 & 0.363 $\pm$ 0.028 & 0.479 $\pm$ 0.027 \\
\hline
\end{tabular}
\end{table}

\subsection{Wall-to-wall crop maps}
\begin{figure}[h!]
  \centering
  \caption{Crop cover maps}
  \includegraphics[width=370pt]{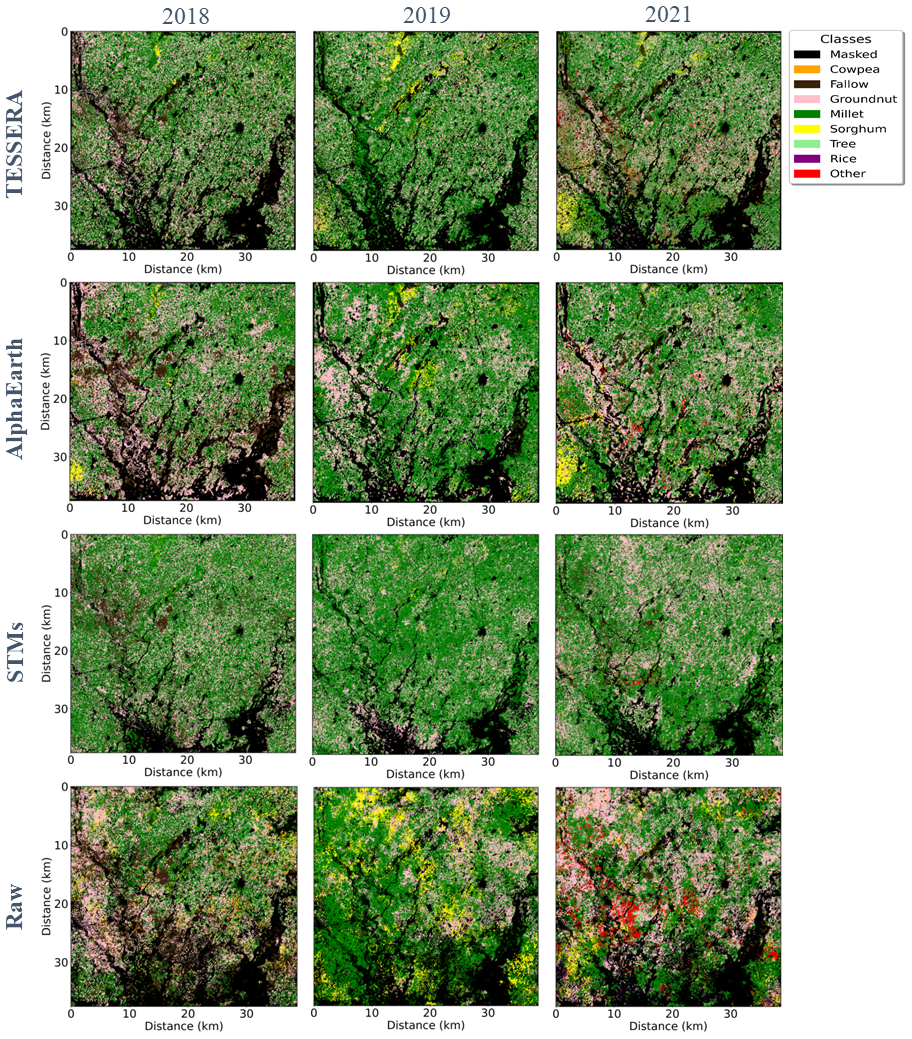}
  \caption{Pictured are the crop cover ensemble maps for TESSERA, AlphaEarth, STMs, and raw input data in 2018, 2019 and 2021. Units on the x and y axis are kilometers of distance. For all input methods, the 2021 map is the most differentiated, similar to the pattern seen in the land cover maps. Of the group, the raw data input maps are the most distinct, showing different patterns, compared to the rest of the group.}
  \label{fig:all_cropcover}
\end{figure}

\begin{figure}[h!]
  \centering
  \caption{Diar\`ere crop type classification map}
  \includegraphics[width=370pt]{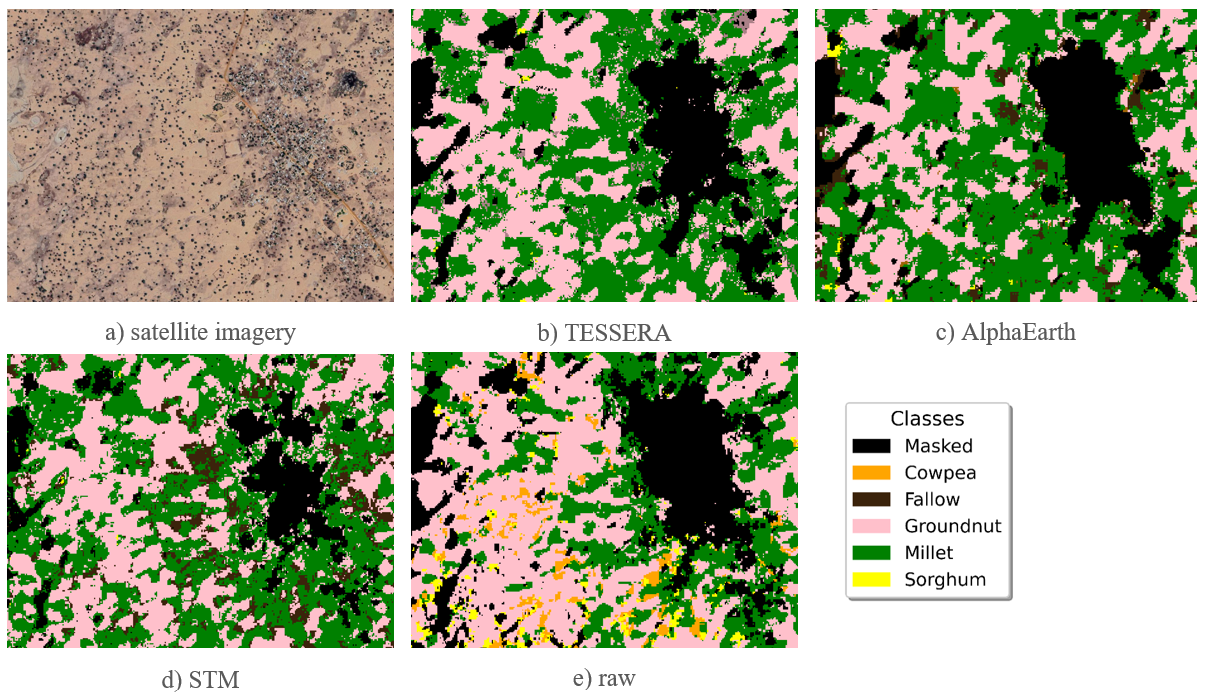}
  \caption{Example crop cover classification around the town of Diar\`ere for a) satellite imagery, b) TESSERA -based classification, c) AlphaEarth -based classification, d) STM -based classification and e) raw -based classification.}
  \label{fig:diarevillage}
\end{figure}

Wall-to-wall crop cover classification maps can be seen in Figure \ref{fig:all_cropcover}. Maps produced by raw data are the least plausible because of large changes between years. In looking at the area around a single town in the center of the study area, Diar\`ere (Figure \ref{fig:diarevillage}), we further observe that TESSERA and AlphaEarth result in relatively similar outputs, where the STM approach appears to under-estimate the built-up area, and the raw approach predicts a high proportion of groundnut. This aligns with the observations on the full-size map and further contextualizes Criterion 2, Plausibility.

\subsection{Temporal transfer learning for \textbf{crop type classification}}

In training on 2018 and evaluating on 2019 for crop mapping, TESSERA offers the highest accuracy and F1 scores (Table \ref{tab:ensemble_results_crop}). There is a 8\% to 10\% decrease in the TESSERA embedding -based performance on each metric compared to both training and testing on 2019, which doesn't account for the evaluation in 2019 using less labels to train due to the withheld evaluation set.  

In training on 2019 and evaluating on 2018, the TESSERA embedding -based approach performance is almost 30\% lower than when training on 2018 and evaluating on 2018. This strengthens the hypothesis that the quality of crop type labels in 2019 is lower than the quality of labels from 2018.  Although the accuracy and F1 metrics are low, the TESSERA embedding -based approach has the highest performance compared to other methods. If a classifier were to predict one of the five crop type classes randomly, we would expect an accuracy of 20\%. In training on 2018, the STM and raw approach perform below 20\%, likewise for STM trained on 2019. This suggests that these methods are worse than a random approach and are not reliable for temporal classifier transfer in this context. In both 2018 and 2019, the AlphaEarth embedding -based approach performs between 9\% and 27\% worse than TESSERA on all metrics.  We do not evaluate for 2021 because, as noted before, there are additional crop classes in the 2021 dataset. 

\begin{table}[h!]
\centering
\small 
\setlength{\tabcolsep}{4pt} 
\renewcommand{\arraystretch}{1.1} 
\resizebox{\columnwidth}{!}{ 
\begin{tabular}{cc l ccc}
\hline
Training Year & Prediction Year & Method & 
\makecell{Accuracy \\ (mean $\pm$ std)} & 
\makecell{Macro avg F1 \\ (mean $\pm$ std)} & 
\makecell{Weighted avg F1 \\ (mean $\pm$ std)} \\
\hline
2018 & 2019 & TESSERA & \textbf{0.626 $\pm$ 0.001} & \textbf{0.325 $\pm$ 0.002} & \textbf{0.601 $\pm$ 0.002} \\
 &  & AlphaEarth & 0.364 $\pm$ 0.002 & 0.180 $\pm$ 0.003 & 0.373 $\pm$ 0.002 \\ 
 &  & STM & 0.172 $\pm$ 0.002 & 0.165 $\pm$ 0.003 & 0.189 $\pm$ 0.002 \\
 &  & Raw & 0.161 $\pm$ 0.003 & 0.155 $\pm$ 0.003 & 0.178 $\pm$ 0.003 \\ 
\hline
2019 & 2018 & TESSERA & \textbf{0.555 $\pm$ 0.001} & \textbf{0.330 $\pm$ 0.001} & \textbf{0.453 $\pm$ 0.002} \\
 &  & AlphaEarth & 0.450 $\pm$ 0.001 & 0.238 $\pm$ 0.002 & 0.355 $\pm$ 0.002 \\ 
 &  & STM & 0.187 $\pm$ 0.002 & 0.147 $\pm$ 0.003 & 0.153 $\pm$ 0.002 \\
 &  & Raw & 0.210 $\pm$ 0.002 & 0.178 $\pm$ 0.002 & 0.210 $\pm$ 0.003 \\ 
\hline
\end{tabular}
}
\caption{Cross-year \textbf{crop classification} performance by training year, prediction year, and method. Values of  $\pm$ are standard deviation over 5 runs, coming from the selection of the validation set. The best results are in \textbf{bold}.}
\label{tab:crossyear_crop_results}
\end{table}

\subsection{Embedding clustering}
We consider a dimensionality reduction of the TESSERA and AlphaEarth embeddings using Uniform Manifold Approximation and Projection (UMAP) \cite{mcinnes_umap_2020} to contextualize the embedding -based approaches. We normalize the embeddings before applying the UMAP with the same parameters for each for comparability. We then color the UMAPs by land cover and crop cover labels.

\begin{figure}[h!]
  \centering
  \includegraphics[width=370pt]{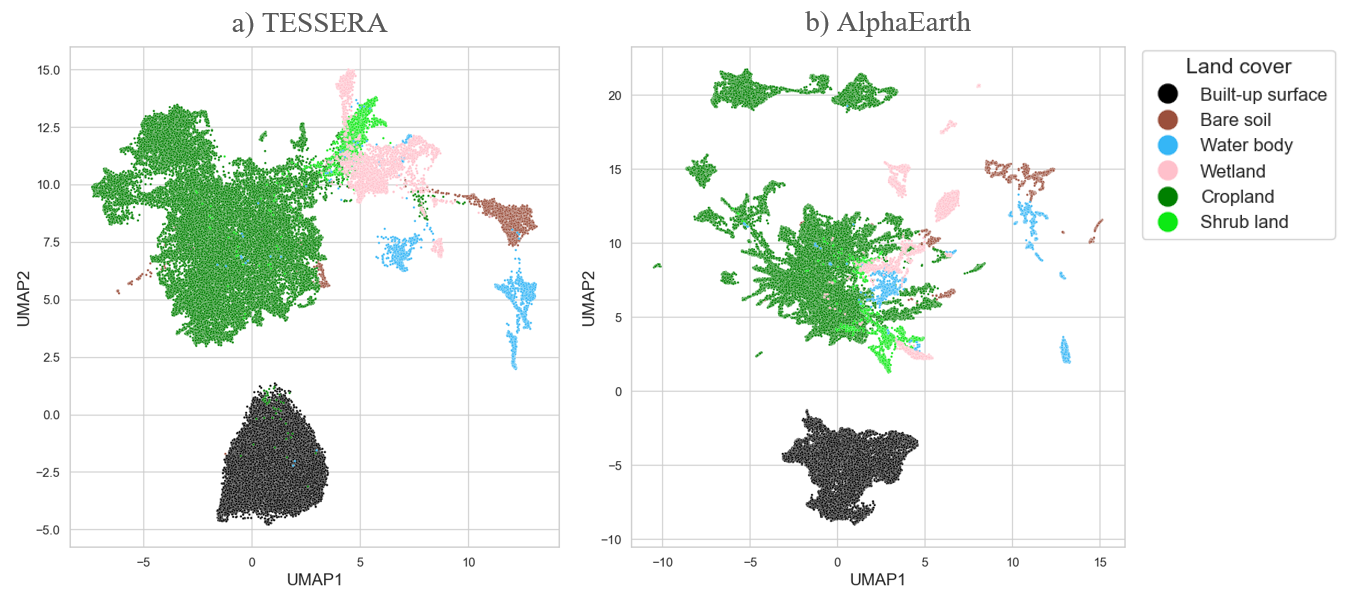}
  \caption{UMAP of a) TESSERA and b) AlphaEarth embeddings of land cover classes for 2018.}
  \label{fig:umaplandcover}
\end{figure}

\begin{figure}[h!]
  \centering
  \includegraphics[width=370pt]{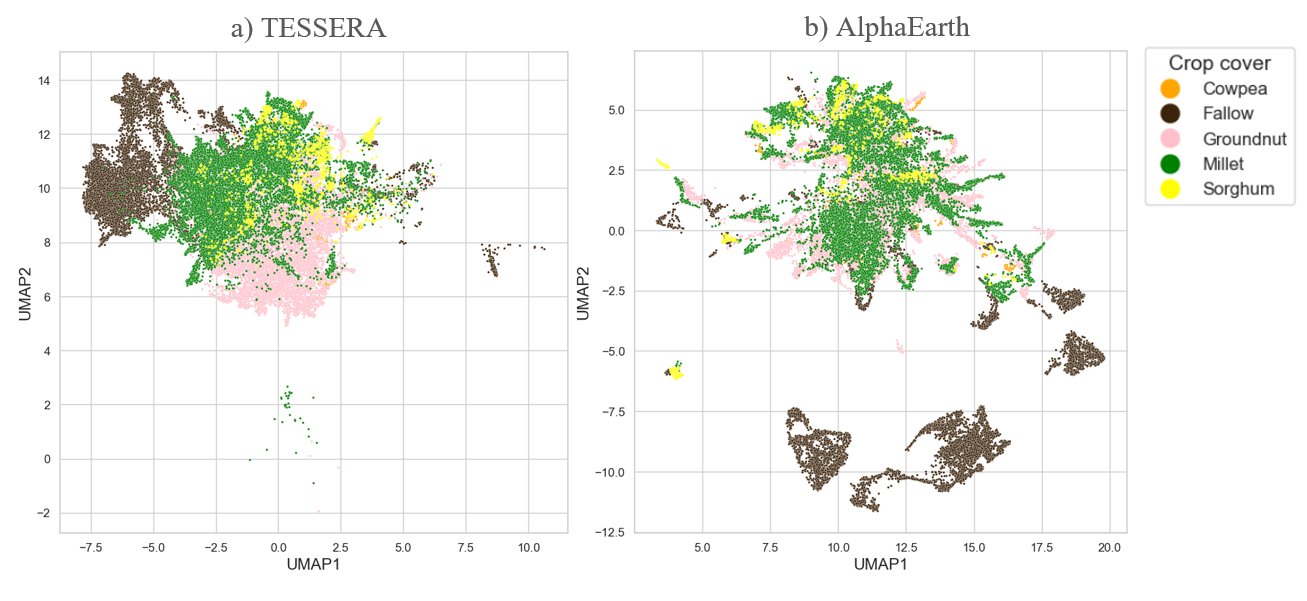}
  \caption{UMAP of a) TESSERA and b) AlphaEarth embeddings of crop cover classes for 2018.}
  \label{fig:umapcropcover}
\end{figure}

When evaluating embeddings from TESSERA and AlphaEarth for land cover classes as seen in Figure \ref{fig:umaplandcover}, we see that both TESSERA and AlphaEarth embeddings clearly separate built-up area labels, although AlphaEarth embeddings seem to spread more in an almost `starburst' shape. For crop cover classes as shown in Figure \ref{fig:umapcropcover}, we find that TESSERA embeddings naturally cluster better, especially the groundnut and fallow groups. The AlphaEarth embeddings again show a `starburst' pattern with varying crops spread around a center. This suggests that TESSERA embeddings have more separable neighborhood structure aligned with the land cover and crop cover semantic groups of interest, which is supported by the land cover and crop type classification results. This explains why TESSERA is better suited for classifying crop type in Senegal.

\subsection{Summary of evaluation}
A summary of performance differences for the 4 criteria (described in Section \ref{sec:criteria}) follows below. In regards to criterion 1, Performance, TESSERA matches or beats other tested methods for both land cover and crop type classification tasks. Evaluation of criterion 2 of Plausibility shows that TESSERA is the strongest, in particular when considering changes to crop land between years. For criterion 3, Transferability, TESSERA shows strong performance in comparison with the other tested approaches, especially in the crop type classification transfer task between years. Finally, for criterion 4, Accessibility, TESSERA and AlphaEarth show strong accessibility for their low compute requirements and elimination for need of feature processing. Across criteria, TESSERA best meets the defined criterion for a valuable crop type classification base.

\section{Discussion}
Our findings demonstrate that embeddings from TESSERA provide a robust and practical approach for crop type classification and mapping in the groundnut basin of Senegal. When compared to AlphaEarth embedding -based approaches and other established methods, TESSERA consistently satisfies the four criteria—Performance, Plausibility, Transferability, and Accessibility—suggesting that it is both technically effective and operationally relevant for smallholder contexts such as in the groundnut basin of Senegal. Further, our work demonstrates that annual TESSERA embeddings offer high classification accuracy and stability across years for the evaluated smallholder agricultural region. 

While we did not not test in other smallholder regions, there is evidence that TESSERA will be effective for crop classification in similar challenging crop classification scenarios. Improvements in temporal transfer learning and in stability between years of classification offer more trustworthy maps to base decision making. Given that smallholder landscapes make up much of the region of rural poor in the world, these improvements have real implications for  food security and poverty reduction for the most vulnerable.

We also find that not all embedding -based methods are suitable for crop type mapping in our smallholder study region in Senegal. While AlphaEarth showed land cover classification metrics matching TESSERA, it performed significantly worse than all tested approaches. We hypothesize that the many data modalities of AlphaEarth as well as the spatial contextual information could `blur' the spectral-temporal signal of crop evolution as well as the spatial boundaries of fields in a way that disadvantages the model in the difficult task of crop type classification. This result supports the idea that choosing a FM based on downstream task will be part of the future of Earth observation. 

Beyond these domain specific advantages, it must also be acknowledged that while pre-generated embeddings from FMs offer significant reductions in compute for downstream tasks, large amounts of compute are needed for the testing, ablation and pretraining processes required in model development. While the paradigm of this work operates assuming that these costs are fixed for the shared benefit of all downstream stakeholders using embeddings, this extra cost cannot be ignored.

Additionally, this work shows evidence that there is varying quality of labels in in the Senegal dataset varies. The resulting variance in accuracy and F1 metrics as well as in the maps supports the need for consistent, high quality labels for crop type mapping. Even with a small number of labels, we report accuracy as high as 84\% (Table \ref{tab:ensemble_results_crop}). However, lower quality labels degrade the performance significantly, as seen for other years of crop classification. We urge future efforts to prioritize consistent data collection strategies between years.

Beyond the scope of this study, several directions offer clear opportunities for further research. Extending evaluations of TESSERA to additional smallholder regions with diverse management systems, crop compositions, and data availability would help assess its global transferability and robustness. Moreover, while our classification captures dominant crop types, it does not explicitly disentangle intercropped or secondary crops—a common and important feature of West African agriculture. Future work could address this by integrating higher-resolution temporal or hyperspectral data to better characterize mixed cropping systems. Finally, systematic error quantification of classification outputs, both spatially and across crop types, would further strengthen confidence in the operational deployment of embedding-based approaches for agricultural monitoring in smallholder landscapes. 

Our work supports the effectiveness of TESSERA embeddings for wall-to-wall land cover and crop type mapping in the groundnut basin of Senegal, and shows promise for broader applications in other smallholder contexts around the globe.

\section{Acknowledgments}
We would like to thank Mantle Labs and UKRI for their support funding this work. 

\bibliographystyle{elsarticle-num} 
\bibliography{171125}







\end{document}